\documentclass{article}
\usepackage{spconf,amsmath,graphicx}
\usepackage{multirow}
\usepackage{amsfonts} 
\usepackage{tikz}
\usepackage{xcolor}

\title{NeRD: Neural field-based Demosaicking}
\name{Tomáš~Kerepecký~$^{1, 2}$\thanks{This work was supported in part by the Czech Science Foundation grant GA21-03921S, the \textit{Praemium Academiae} awarded by the Czech Academy of Sciences, and the Fulbright commission under the Fulbright-Masaryk award.}
, Filip Šroubek~$^{1}$
, Adam Novozámský~$^{1}$
, Jan Flusser~$^{1}$}
\address{
{\normalsize $^{1}$ Institute of Information Theory and Automation, The Czech Academy of Sciences, Czechia} \\
{\normalsize $^{2}$Faculty of Nuclear Sciences and Physical Engineering, Czech Technical University in Prague, Czechia} \\
}

\begin{document}
%
\maketitle
\begin{abstract}
We introduce NeRD, a new demosaicking method for generating full-color images from Bayer patterns. Our approach leverages advancements in neural fields to perform demosaicking by representing an image as a coordinate-based neural network with sine activation functions. The inputs to the network are spatial coordinates and a low-resolution Bayer pattern, while the outputs are the corresponding RGB values. An encoder network, which is a blend of ResNet and U-net, enhances the implicit neural representation of the image to improve its quality and ensure spatial consistency through prior learning. Our experimental results demonstrate that NeRD outperforms traditional and state-of-the-art CNN-based methods and significantly closes the gap to transformer-based methods.
\end{abstract}
\section{Introduction}
\label{sec:intro}

\begin{figure}[!t]
\centering
\includegraphics[width=0.45\textwidth]{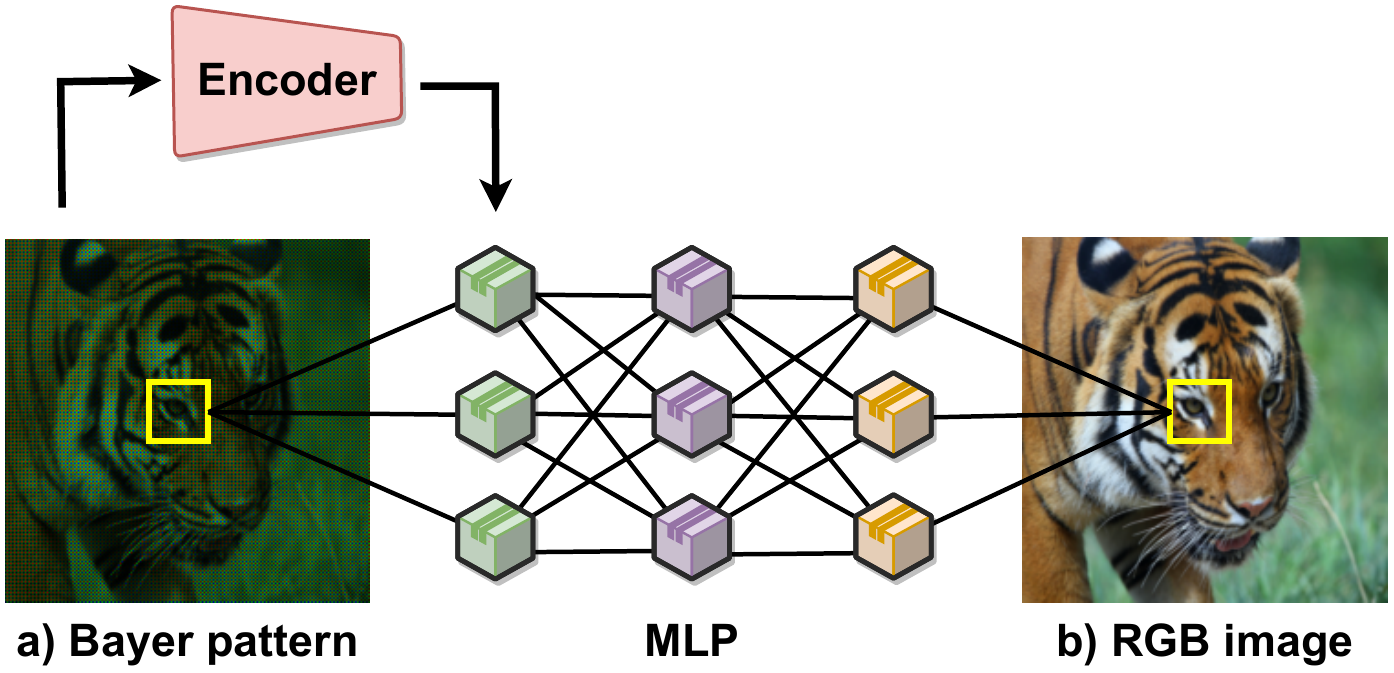}%
\caption{An illustration of demosaicking using coordinate-based Multilayer Perceptron and local encoding technique.\label{fig:header}}
\end{figure}

Raw data acquired by modern digital camera sensors is subject to various types of signal degradation, one of the most severe being the color filter array. To convert the raw data (Fig.~\ref{fig:header}a) into an image suitable for human visual perception (Fig.~\ref{fig:header}b), a demosaicking procedure is necessary \cite{menon2011color}.

Two main categories of image demosaicking exist: model-based and learning-based methods. Model-based methods, such as bilinear interpolation, Malvar \cite{malvar2004high}, or Menon \cite{menon2006demosaicing}, are still widely used, but they fail to match the performance of recent deep learning-based approaches using deep convolutional networks (CNN) \cite{kokkinos2019iterative, gharbi2016deep, kerepecky2021d3net} or Swin Transformers \cite{xing2022residual}.

Recently, Transformer networks have seen remarkable success in computer vision tasks and have become a state-of-the-art approach in demosaicking. However, a new paradigm in deep learning, Neural Fields (NF) \cite{xie2022neural}, is gaining attention due to its comparable or superior performance in several computer vision tasks \cite{xie2022neural, mildenhall2021nerf, shangguan2022learning, chen2021nerv, chen2021learning, sitzmann2020implicit, skorokhodov2021adversarial}. The basic idea behind NF is to represent data as the weights of a Multilayer Perceptron (MLP), known as implicit neural representation.

NF has been applied in various domains and applications including Neural Radiance Fields (NeRF) \cite{mildenhall2021nerf} which achieved state-of-the-art results in representing complex 3D scenes. NeRV \cite{chen2021nerv} encodes entire videos in neural networks. The Local Implicit Image Function (LIIF) \cite{chen2021learning} represents an image as a neural field capable of extrapolating to 30 times higher resolution. SIREN \cite{sitzmann2020implicit} uses a sinusoidal neural representation and demonstrates superiority over classical ReLU MLP in representing complex natural signals such as images.

Prior information from training data can be encoded into neural representation through conditioning (local or global) using methods such as concatenation, modulation of activation functions \cite{dupont2022data}, or hypernetworks \cite{skorokhodov2021adversarial}. For example, CURE \cite{shangguan2022learning}, a state-of-the-art method for video interpolation based on NF, uses an encoder to impose space-time consistency using local feature codes.

NF has also been used in image-to-image translation tasks such as  superresolution, denoising, inpainting, and generative modeling \cite{xie2022neural}. However, to the best of our knowledge, no NF method has been proposed for demosaicking.

In this paper, we present NeRD, a novel approach for image demosaicking based on NF. The proposed method employs a joint ResNet and U-Net architecture to extract prior information from high-resolution ground-truth images and their corresponding Bayer patterns. This information is then used to condition the MLP using local feature encodings. The proposed approach offers a unique and innovative solution for image demosaicking.

\begin{figure*}[!t]
  \includegraphics[width=\textwidth]{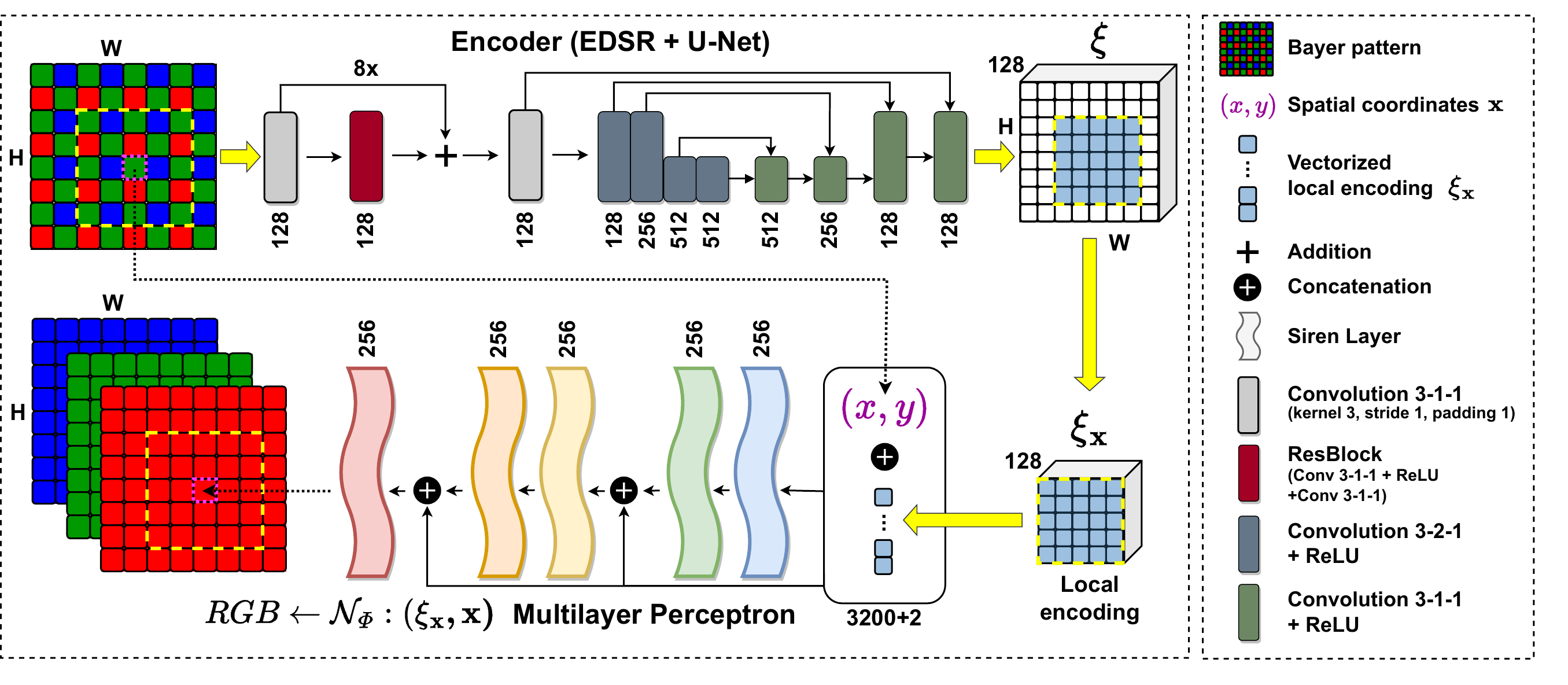}
  \caption{The overall architecture of NeRD.  Encoder consisting of 8 residual blocks and U-net architecture generates encoding $\xi$ for the input Bayer pattern.  Numbers below each layer in the encoder represent the number of output channels. Spatial coordinates $ {\bf x} = (x,y) $ concatenated with the corresponding local encoding vector $\xi_{\bf x}$ are transformed into RGB value using a multilayer perceptron with 5 hidden layers each with 256 output channels, siren activation functions, and two skip connections. \label{fig:nerd}}
\end{figure*}

\newpage

\section{Proposed method}
\label{sec:propo}

NeRD converts spatial coordinates and local encodings into RGB values. The local encodings are generated by an encoder that integrates consistency priors in NeRD. The overall architecture of NeRD is depicted in Fig.~\ref{fig:nerd}.

The core of NeRD is a fully connected feedforward network $\mathcal{N}_\Phi: \left(\xi_{\bf x}, {\bf x} \right) \rightarrow {\bf n}$ with 5 hidden layers, each with 256 output channels and sine activation functions. $\Phi$ denotes the network weights. The input is a spatial coordinate $ {\bf x}~=~(x,y)~\in~\mathbb{R}^2$ and local encoding vector $\xi_{\bf x}$. The output is a single RGB value ${\bf n} = (r,g,b) \in \mathbb{R}^3$. The SIREN architecture~\cite{sitzmann2020implicit} was chosen for its ability to model signals with greater precision compared to MLPs with ReLU. There are two skip connections that concatenate the input vector with the output of the second and fourth hidden layers.

\begin{figure*}[t!]
\centering
\includegraphics[width=0.95\textwidth]{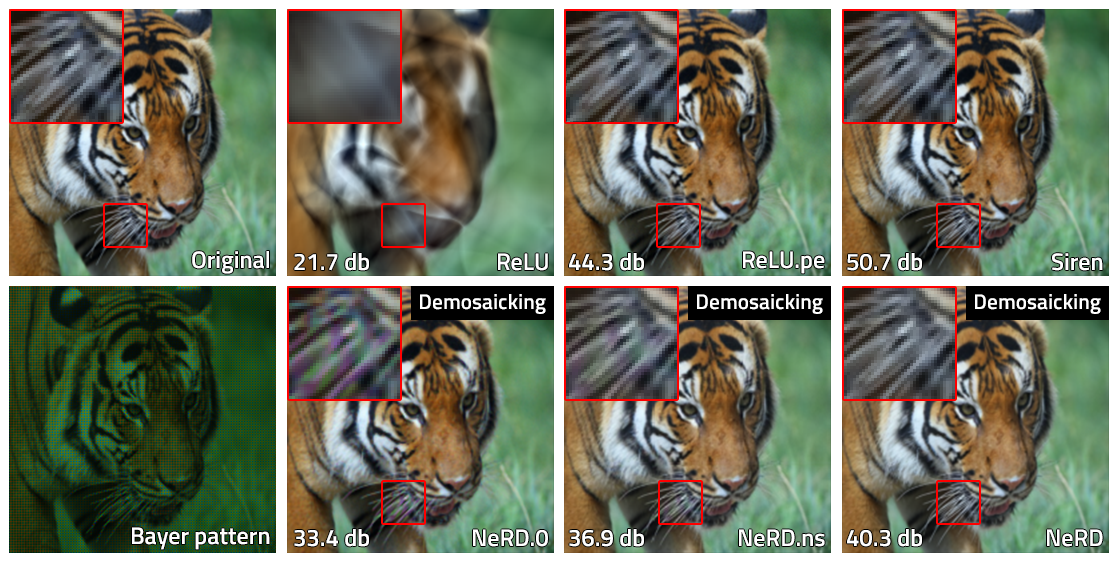}
\caption{The ablation study of NeRD. The original image is from DIV2K dataset. "ReLU" and "Siren" models show the implicit neural representation of the original image using MLP with ReLU and sine activation functions, respectively.  These models were trained in a self-supervised manner to fit the original image. "ReLU.pe" stands for "ReLU" model with additional positional encoding in the form of Fourier feature mapping. "NeRD.0" model is identical to  "Siren" model but is only trained using the input Bayer pattern. "NeRD" is the proposed demosaicking method, while "NeRD.ns" represents the proposed architecture without skip connections in the MLP. Each image is labeled with its PSNR value with respect to the original image. \label{fig:ablation}}
\end{figure*}

Using the MLP without local encoding $\xi_{\bf x}$ leads to suboptimal demosaicking results due to the insufficient information contained in the training image. This is demonstrated by the result in Fig.~\ref{fig:ablation}-NeRD.0, where the reconstructed image is the output of the SIREN model trained only on original input Bayer pattern  in self-supervised manner. The lack of spatial consistency in these results highlights the need for additional prior information in the form of spatial encoding, which is why we utilize an encoder.

The encoder provides local feature codes $\xi_{\bf x}$ for a given coordinate ${\bf x}$ and its architecture is shown in the first row of Fig.~\ref{fig:nerd}. The Bayer pattern is processed through a combined network that incorporates 8 residual blocks (using the EDSR architecture~\cite{lim2017enhanced}) and 4 downsampling and 4 upsampling layers (U-Net architecture \cite{ronneberger2015u}) connected by multiple skip connections. The result is a global feature encoding $H \times W \times 128$, where $H$ and $W$ denote the height and width of the initial Bayer pattern in pixels. The local encoding $\xi_{\bf{x}}$ is extracted from the global encoding as a $5\times5$ region centered at $\bf{x}$, which is then flattened into a 3200-dimensional feature vector. The architecture of the encoder is adopted from \cite{shangguan2022learning}. 

The final RGB image is produced by independently retrieving the RGB pixel values from NeRD at the coordinates specified by the input Bayer pattern.

\section{Experiment}
\label{sec:exper}

We numerically validated NeRD on standard image datasets. Experiments also include an ablation study highlighting the key components of the proposed architecture and comparisons with state-of-the-art methods.

\subsection{Dataset and Evaluation Metrics}
\label{ssec:data}

A training set was created by combining multiple high-resolution datasets, such as DIV2K \cite{Agustsson_2017_CVPR_Workshops}, Flickr2K \cite{lim2017enhanced}, and OST \cite{wang2018sftgan}, resulting in a total of 12\,000 images. During each epoch, 10\,000 randomly cropped patches of size $200 \times 200$ and corresponding Bayer patterns (GBRG) were generated. The Kodak and McM \cite{zhang2011color} datasets were used for testing. The~evaluation was performed using  Peak Signal to Noise Ratio (PSNR) and the Structural Similarity Index Measure (SSIM).

\subsection{Training Configuration}
\label{ssec:subhead1}

The training was conducted using an Nvidia A100 GPU. The NeRD model was optimized using the Mean Squared Error loss function, and the Adam optimizer was used with $\beta_1$ = 0.9 and $\beta_2$ = 0.999. The initial learning rate was set to 0.0001, and a step decay was applied, reducing the learning rate by 0.95 every epoch consisting of 10\,000 iterations. The patch size was set to $200 \times 200$ and the batch size was 5.

\subsection{Ablation Study}
\label{ssec:subhead2}

\hspace{15px} \textbf{MLP and activation functions.} RGB images can be represented as the weights of a fully connected feedforward neural network. This representation is achieved by training an MLP in a self-supervised manner to fit the original image. However, the usage of standard ReLU activation functions in MLPs produces unsatisfactory results, as shown in~\mbox{Fig. \ref{fig:ablation}-ReLU}. To significantly improve reconstruction, Fourier feature mapping of input spatial coordinates can be used (see~Fig.~\ref{fig:ablation}-ReLU.pe). This technique is referred to as “positional encoding”. Nonetheless, an even better outcome can be achieved by replacing ReLU with sine functions, also known as SIRENs. They demonstrate the capability of MLPs as image decoders and hold promise for demosaicking applications. SIREN architecture has the capacity to model RGB images with great precision. As demonstrated in Fig.~\ref{fig:ablation}-Siren, the~SIREN with 5 hidden layers, each with 256 neurons, achieved a PSNR of 50.7 dB when trained for just 1000 iterations to fit the original image.

\textbf{Encoder.} The naive approach of decoding RGB images from Bayer patterns using SIREN architecture fails as it loses two-thirds of the original information, as shown in Fig.~\ref{fig:ablation}-NeRD.0. To improve the demosaicking capability of the MLP, prior information must be incorporated through an encoder. This encoder learns prior information across various training image pairs and conditions the MLP with local encodings. The effectiveness of the encoder is demonstrated in Fig.~\ref{fig:ablation}-NeRD, which shows the results of demosaicking using the  NeRD architecture described in Sec. \ref{sec:propo}.

\begin{figure*}[t!]
\centering
\begingroup
\setlength{\tabcolsep}{1pt}
\begin{tabular}{cccccc}
\multicolumn{2}{c}{\multirow[c]{3}{*}[.108\textwidth]{\includegraphics[width=0.36\textwidth]{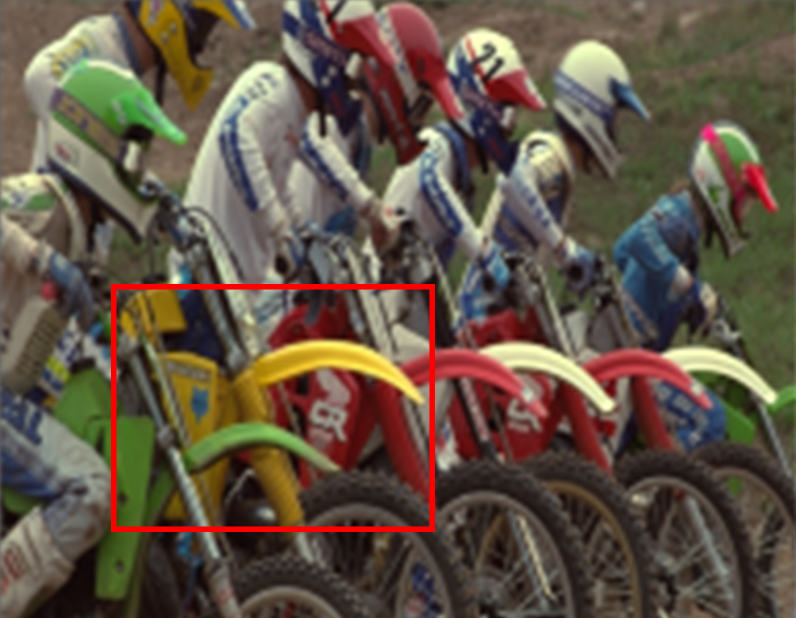}}}
& \includegraphics[width=0.16\textwidth]{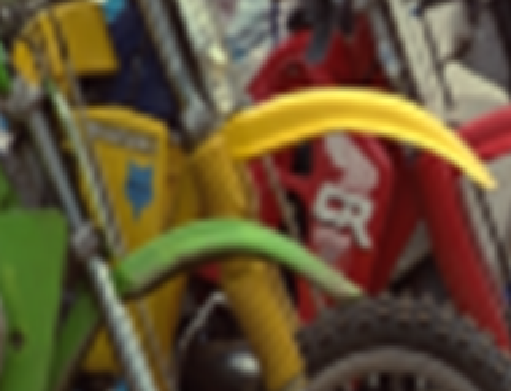}
& \includegraphics[width=0.16\textwidth]{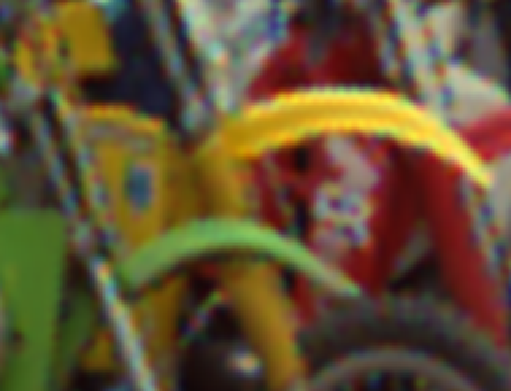}
& \includegraphics[width=0.16\textwidth]{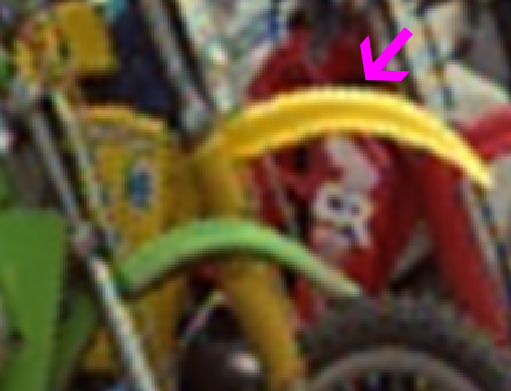}%
& \includegraphics[width=0.16\textwidth]{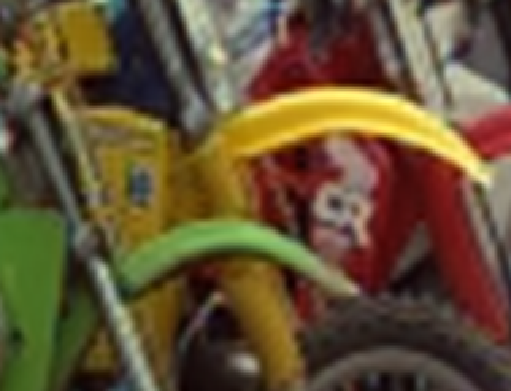}\\
& & (b) Ground-truth & (c) Bilinear & (d) Matlab & (e) Menon  \\
& & \includegraphics[width=0.16\textwidth]{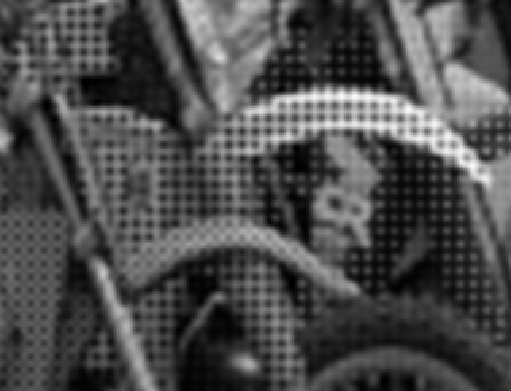}
& \includegraphics[width=0.16\textwidth]{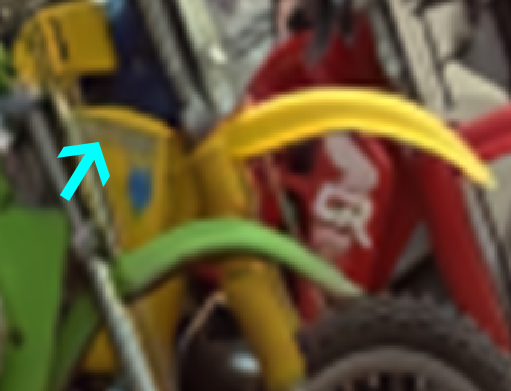}
& \includegraphics[width=0.16\textwidth]{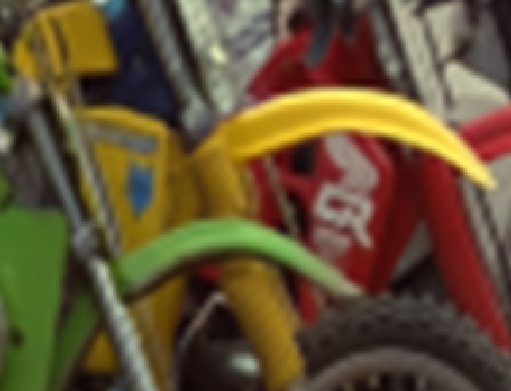}%
& \includegraphics[width=0.16\textwidth]{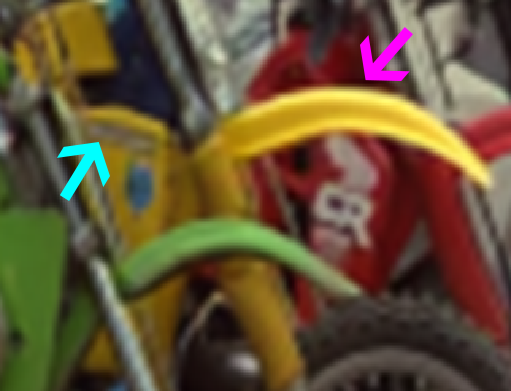}\\
\multicolumn{2}{c}{(a) Kodak:kodim05.png (cropped)}  & (f) Raw & (g) DeepDemosaick   & (h) RSTCANet  
& (i) NeRD\\
\end{tabular}
\caption{A visual comparison of NeRD and the current state-of-the-art methods on an example from the Kodak dataset. The visual differences are highlighted by close-ups, which correspond to the red box in the original image. Although NeRD exhibits slightly inferior visual performance compared to RSTCANet, it outperforms traditional methods in terms of reconstruction  accuracy (indicated by the magenta arrow) and avoids over-smoothing details, as seen with the DeepDemosaick  method (indicated by the cyan arrow). \label{fig:compare} }
\endgroup

\end{figure*}

\textbf{Skip Connections.} The integration of encoding into the MLP can be achieved through various methods. However, methods such as modulation of activation functions or the use of hypernetworks present challenges in terms of parallelization. Hence, we utilized a method of concatenation, where the coordinates and feature vectors are combined at the input and later concatenation of the input with the second and fourth hidden layers is performed using skip connections. The significance of incorporating skip connections into the MLP is illustrated in Fig.~\ref{fig:ablation}-NeRD.ns (no-skip). This figure demonstrates a degradation in both the quality of the reconstruction and the PSNR value when these connections are omitted.

\begin{table}[b!]
  \caption{Average PSNR/SSIM obtained by NeRD and the current state-of-the-art methods on the McM and Kodak datasets. 
\textbf{Bold} and \underline{underline} highlights the highest and second highest values, respectively. Note
the superior results of NeRD over the CNN-based and traditional methods. Only RSTCANet, which is based on transformers, has slightly higher scores.  \newline}
  	\centering
  	\resizebox{0.95\columnwidth}{!}{\begin{tabular}{ccc}
	\hline\hline
  \multirow{2}{*}{Method} & McM \cite{zhang2011color} & Kodak \\ 
  & PSNR/SSIM & PSNR/SSIM \\ 
 	[0.5ex]
 	\hline 
Bilinear                                                  &  27.15/0.912  &  28.01/0.894  \\
Matlab (Malvar) \cite{malvar2004high}                     &  30.54/0.923  &  33.52/0.957 \\
Menon  \cite{menon2006demosaicing}                        &  31.40/0.918  &  35.20/0.968 \\
DeepDemosaick   \cite{kokkinos2019iterative}              &  33.31/0.942  &  37.76/0.976 \\
RSTCANet        \cite{xing2022residual}                   &  \textbf{37.77/0.978}  & \textbf{ 40.84/0.988}\\ 
NeRD                                                      &  \underline{36.18/0.969}  &  \underline{39.07/0.984} \\
\hline
	\end{tabular}}
	\label{table:compare}
\end{table}

\subsection{Comparison With Existing Methods}
\label{ssec:subhead3}

The evaluation of the proposed NeRD demosaicking algorithm was performed on the McM and Kodak datasets, which were resized and cropped to $200\times200$ px. A comparison of NeRD with traditional demosaicking algorithms and state-of-the-art methods is presented in Table \ref{table:compare} in terms of average PSNR and SSIM values calculated from the demosaicked images. The results show that NeRD outperforms traditional methods and the CNN-based DeepDemosaick \cite{kokkinos2019iterative}, but falls slightly behind the transformer-based RSTCANet \cite{xing2022residual}. 

A visual comparison of the demosaicked images is presented in Fig. \ref{fig:compare}. The figure highlights differences between NeRD and the other methods and provides insights into their performance. One notable characteristic of NeRD is that it avoids over-smoothing details, unlike the DeepDemosaick~\cite{kokkinos2019iterative} method, as indicated by the cyan arrow in the~Fig.~\ref{fig:compare}g. Furthermore, NeRD outperforms traditional methods in terms of preserving fine details and avoiding  unpleasant artifacts, as indicated by the magenta arrow in the Fig.~\ref{fig:compare}d.

\section{Conclusion}

This paper presents a novel demosaicking algorithm, NeRD, that leverages the recent class of techniques known as Neural Fields. The ablation study results emphasize the significance of incorporating an encoder and skip connections within the MLP, which results in significant improvement over traditional techniques and outperforms the CNN-based DeepDemosaick method in preserving fine details while avoiding undesirable artifacts. Although NeRD shows slightly lower visual performance compared to the transformer-based \mbox{RSTCANet}, it still demonstrates remarkable accuracy in terms of reconstruction. Future research can focus on enhancing NeRD through fine-tuning using input Bayer pattern-specific loss functions and integrating Transformer networks or \mbox{ConvNeXt} into the encoder. In addition, expanding the training set by more diverse datasets can improve the prior. Albeit NeRD may not attain the performance level of Transformer-based demosaicking, our contribution broadens the range of domains where Neural Fields can be applied.

\vfill\pagebreak

\bibliographystyle{IEEEbib}
\bibliography{refs}

\begin{thebibliography}{10}

\bibitem{menon2011color}
Daniele Menon and Giancarlo Calvagno,
\newblock ``Color image demosaicking: An overview,''
\newblock {\em Signal Processing: Image Communication}, vol. 26, no. 8-9, pp.
  518--533, 2011.

\bibitem{malvar2004high}
Henrique~S Malvar, Li-wei He, and Ross Cutler,
\newblock ``High-quality linear interpolation for demosaicing of
  bayer-patterned color images,''
\newblock in {\em 2004 IEEE International Conference on Acoustics, Speech, and
  Signal Processing}. IEEE, 2004, vol.~3, pp. iii--485.

\bibitem{menon2006demosaicing}
Daniele Menon, Stefano Andriani, and Giancarlo Calvagno,
\newblock ``Demosaicing with directional filtering and a posteriori decision,''
\newblock {\em IEEE Transactions on Image Processing}, vol. 16, no. 1, pp.
  132--141, 2006.

\bibitem{kokkinos2019iterative}
Filippos Kokkinos and Stamatios Lefkimmiatis,
\newblock ``Iterative joint image demosaicking and denoising using a residual
  denoising network,''
\newblock {\em IEEE Transactions on Image Processing}, vol. 28, no. 8, pp.
  4177--4188, 2019.

\bibitem{gharbi2016deep}
Micha{\"e}l Gharbi, Gaurav Chaurasia, Sylvain Paris, and Fr{\'e}do Durand,
\newblock ``Deep joint demosaicking and denoising,''
\newblock {\em ACM Transactions on Graphics (ToG)}, vol. 35, no. 6, pp. 1--12,
  2016.

\bibitem{kerepecky2021d3net}
Tom{\'a}{\v{s}} Kerepecky and Filip {\v{S}}roubek,
\newblock ``D3net: Joint demosaicking, deblurring and deringing,''
\newblock in {\em 2020 25th International Conference on Pattern Recognition
  (ICPR)}. IEEE, 2021, pp. 1--8.

\bibitem{xing2022residual}
Wenzhu Xing and Karen Egiazarian,
\newblock ``Residual swin transformer channel attention network for image
  demosaicing,''
\newblock in {\em 2022 10th European Workshop on Visual Information Processing
  (EUVIP)}. IEEE, 2022, pp. 1--6.

\bibitem{xie2022neural}
Yiheng Xie, Towaki Takikawa, Shunsuke Saito, Or~Litany, Shiqin Yan, Numair
  Khan, Federico Tombari, James Tompkin, Vincent Sitzmann, and Srinath Sridhar,
\newblock ``Neural fields in visual computing and beyond,''
\newblock in {\em Computer Graphics Forum}. Wiley Online Library, 2022,
  vol.~41, pp. 641--676.

\bibitem{mildenhall2021nerf}
Ben Mildenhall, Pratul~P Srinivasan, Matthew Tancik, Jonathan~T Barron, Ravi
  Ramamoorthi, and Ren Ng,
\newblock ``Nerf: Representing scenes as neural radiance fields for view
  synthesis,''
\newblock {\em Communications of the ACM}, vol. 65, no. 1, pp. 99--106, 2021.

\bibitem{shangguan2022learning}
Wentao Shangguan, Yu~Sun, Weijie Gan, and Ulugbek~S Kamilov,
\newblock ``Learning cross-video neural representations for high-quality frame
  interpolation,''
\newblock in {\em Computer Vision--ECCV 2022: 17th European Conference, Tel
  Aviv, Israel, October 23--27, 2022, Proceedings, Part XV}. Springer, 2022,
  pp. 511--528.

\bibitem{chen2021nerv}
Hao Chen, Bo~He, Hanyu Wang, Yixuan Ren, Ser~Nam Lim, and Abhinav Shrivastava,
\newblock ``Nerv: Neural representations for videos,''
\newblock {\em Advances in Neural Information Processing Systems}, vol. 34, pp.
  21557--21568, 2021.

\bibitem{chen2021learning}
Yinbo Chen, Sifei Liu, and Xiaolong Wang,
\newblock ``Learning continuous image representation with local implicit image
  function,''
\newblock in {\em Proceedings of the IEEE/CVF Conference on Computer Vision and
  Pattern Recognition}, 2021, pp. 8628--8638.

\bibitem{sitzmann2020implicit}
Vincent Sitzmann, Julien Martel, Alexander Bergman, David Lindell, and Gordon
  Wetzstein,
\newblock ``Implicit neural representations with periodic activation
  functions,''
\newblock {\em Advances in Neural Information Processing Systems}, vol. 33, pp.
  7462--7473, 2020.

\bibitem{skorokhodov2021adversarial}
Ivan Skorokhodov, Savva Ignatyev, and Mohamed Elhoseiny,
\newblock ``Adversarial generation of continuous images,''
\newblock in {\em Proceedings of the IEEE/CVF Conference on Computer Vision and
  Pattern Recognition}, 2021, pp. 10753--10764.

\bibitem{dupont2022data}
Emilien Dupont, Hyunjik Kim, SM~Eslami, Danilo Rezende, and Dan Rosenbaum,
\newblock ``From data to functa: Your data point is a function and you should
  treat it like one,''
\newblock {\em arXiv preprint arXiv:2201.12204}, 2022.

\bibitem{lim2017enhanced}
Bee Lim, Sanghyun Son, Heewon Kim, Seungjun Nah, and Kyoung Mu~Lee,
\newblock ``Enhanced deep residual networks for single image
  super-resolution,''
\newblock in {\em Proceedings of the IEEE conference on computer vision and
  pattern recognition workshops}, 2017, pp. 136--144.

\bibitem{ronneberger2015u}
Olaf Ronneberger, Philipp Fischer, and Thomas Brox,
\newblock ``U-net: Convolutional networks for biomedical image segmentation,''
\newblock in {\em Medical Image Computing and Computer-Assisted
  Intervention--MICCAI 2015: 18th International Conference, Munich, Germany,
  October 5-9, 2015, Proceedings, Part III 18}. Springer, 2015, pp. 234--241.

\bibitem{Agustsson_2017_CVPR_Workshops}
Eirikur Agustsson and Radu Timofte,
\newblock ``Ntire 2017 challenge on single image super-resolution: Dataset and
  study,''
\newblock in {\em The IEEE Conference on Computer Vision and Pattern
  Recognition (CVPR) Workshops}, July 2017.

\bibitem{wang2018sftgan}
Chao~Dong Xintao~Wang, Ke~Yu and Chen~Change Loy,
\newblock ``Recovering realistic texture in image super-resolution by deep
  spatial feature transform,''
\newblock in {\em IEEE Conference on Computer Vision and Pattern Recognition
  (CVPR)}, 2018.

\bibitem{zhang2011color}
Lei Zhang, Xiaolin Wu, Antoni Buades, and Xin Li,
\newblock ``Color demosaicking by local directional interpolation and nonlocal
  adaptive thresholding,''
\newblock {\em Journal of Electronic imaging}, vol. 20, no. 2, pp.
  023016--023016, 2011.

\end{thebibliography}

\end{document}